\documentclass{sig-alternate}

\usepackage{amsmath} 
\usepackage{amssymb} 
\usepackage{subcaption,graphicx,caption,subfloat}

\usepackage{pgfplots}\pgfplotsset{compat=1.4} 

\graphicspath{{./}{figures/}}

 \newtheorem{lemma}{Lemma}[section]
\newtheorem{theorem}[lemma]{Theorem}
 \newtheorem{corollary}[lemma]{Corollary}
 \newtheorem{definition}[lemma]{Definition}

\newcommand{\R}[1]{{\rm #1}} \newcommand{\mB}[1]{{\mathbb{#1}}}

\newcommand{\half}{\mbox{\small$\frac{1}{2}$}}
\DeclareMathOperator*{\argmin}{argmin}

\newcommand{\Snp}{\mathcal{S}^n_+}

\begin{document}
%
\conferenceinfo{KDD}{2014 New York City, NY USA}

\title{Beyond L2-Loss Functions for Learning Sparse Models}
%

\numberofauthors{3}
\author{
\alignauthor
Karthikeyan Natesan Ramamurthy\\
       \affaddr{IBM Thomas J. Watson Reseach Center}\\
       \affaddr{1101 Kitchawan Road}\\
       \affaddr{Yorktown Heights, NY, USA}\\
       \email{knatesa@us.ibm.com}
\alignauthor
Aleksandr Y. Aravkin\\
       \affaddr{IBM Thomas J. Watson Reseach Center}\\
       \affaddr{1101 Kitchawan Road}\\
       \affaddr{Yorktown Heights, NY, USA}\\
       \email{saravkin@us.ibm.com}
\and
\alignauthor 
Jayaraman J. Thiagarajan \\
       \affaddr{Lawrence Livermore National Laboratory}\\
       \affaddr{7000 East Ave}\\
       \affaddr{Livermore, CA, USA}\\
       \email{jayaramanthi1@llnl.gov}
}

\maketitle
\begin{abstract}
Incorporating sparsity priors in learning tasks can give rise to simple, and interpretable models for complex high dimensional data. 
Sparse models have found widespread use in structure discovery, recovering data from corruptions, and a variety of large scale unsupervised and supervised learning problems. 
Assuming the availability of sufficient data, these methods infer dictionaries for sparse representations by optimizing for high-fidelity reconstruction. 
In most scenarios, the reconstruction quality is measured using the squared Euclidean distance, and efficient algorithms have  been developed for both batch and online learning cases. 
However, new application domains motivate 
looking beyond conventional loss functions. 
For example, robust loss functions such as $\ell_1$ and Huber are useful in learning outlier-resilient models, 
and the quantile loss is beneficial in discovering structures that are the representative of a particular quantile.  
These new applications motivate our work in generalizing sparse learning to a broad class of convex loss functions. 
In particular, we consider the class of piecewise linear quadratic (PLQ) cost functions that includes Huber, as well as $\ell_1$,  
quantile, Vapnik, hinge loss, and smoothed variants of these penalties. 
We propose an algorithm to learn dictionaries and obtain sparse codes when the data reconstruction fidelity is measured using any smooth PLQ cost function.  
We provide convergence guarantees for the proposed algorithm, and demonstrate the convergence behavior using empirical experiments. 
Furthermore, we present three case studies that require the use of PLQ cost functions:
 (i) robust image modeling,
 (ii) tag refinement for image annotation and retrieval and 
 (iii) computing empirical confidence limits for subspace clustering.
\end{abstract}

%


\section{Introduction}
\label{sec:intro}
Deriving predictive inference from data requires both approximating the generating process using a model,  
and estimating model parameters from input data and the observed responses. 
The generating process can be approximated as $y_i \approx f(x_i;a)$ where $x_i \in \mathbb{R}^{K}$ is the input data sample, 
$y_i \in \mathbb{R}$ is the corresponding response, 
$f$ is the assumed model and $a \in \mathbb{R}^K$ are the parameters. 
When $f$ is linear, this representation reduces to the classical linear model $y_i \approx x_i^T a$. 
Given the vector of observed responses $y \in \mathbb{R}^{T}$, and input data matrix $X  = [x_1 x_2 \ldots x_T] \in \mathbb{R}^{K \times T}$, the parameters $a$ can be estimated using linear regression, where the $\ell_2$ loss on the residual $r = y-X^T a$ is minimized. The complexity of the linear model can be reduced by shrinking the small entries in $a$ to zero \cite{hastie2009elements}. 
This approach gives a sparse linear model, where only a small fraction of the parameters are non-zero (and hence active). 
Sparse parameters allow improved model interpretability because of parsimony \cite{}. 
From the viewpoint of statistical learning theory, sparsity also improves the generalizability, and hence usefulness of the model.   

When the observations are high-dimensional, 
we can denote each observation vector as $y \in \mathbb{R}^M$, 
and assume that it can be approximated using a sparse linear combination of representative columns in the \textit{dictionary} matrix $D \in \mathbb{R}^{M \times K}$. 
The parameter vector $a$, also referred to as the {\it sparse code} of $y$, can be obtained by solving an optimization problem of the form 
\begin{equation}
\min_{a} \rho_1(y-Da) + \lambda \rho_2(a)
\label{eqn:sparse_coding}
\end{equation} where $\rho_1$ is the \textit{loss function} that measures the distance between $y$ and $Da$, 
$\rho_2$ is the sparsity \textit{regularizer} on $a$, 
and $\lambda$ is the regularization penalty that controls the trade-off between loss and regularization. The choice of loss function $\rho_1$ corresponds to the noise or deviation model for the discrepancy between the observed and predicted data. Sparse models have had widespread applications in speech and audio processing \cite{giacobello2012sparse,sivaram2010sparse,gemmeke2011exemplar}, image analysis and recovery \cite{taswell2000and,Elad_book,starck2010sparse}, compressive sampling \cite{donoho2006compressed}, blind source separation \cite{zibulevsky2001blind,li2004analysis,gribonval2006survey}, unsupervised \cite{Ramirez,l1graph}, supervised \cite{Aviyente2006,wright,scspm,mairal2008supervised}, semi-supervised \cite{yan2009semi} and transfer learning \cite{raina2007self,maurer2012sparse}.

So far, we have assumed that a pre-defined dictionary $D$ is available for sparse coding. However, given a set of $T$ observations, $\{y_i\}_{i=1}^T$, where $T$ is sufficiently large, 
the dictionary $D$ can be adapted from the data itself, by jointly minimizing the sum of $T$ objectives in (\ref{eqn:sparse_coding}) over $D$ and $\{a\}_{i=1}^T$.
Additional constraints may also be placed on the dictionary and the sparse codes. 
Most of the existing dictionary learning frameworks in the literature \cite{Elad_KSVD,Engan1999,Engan1999_1,rubinstein2010dictionaries,tosic2011dictionary,mairal2010online} are customized to the case where $\rho_1$ is the $\ell_2$ loss function. Some applications where dictionaries inferred with the $\ell_1$ misfit loss have been effective include robust background modeling \cite{sivalingam2011dictionary}, emerging topic detection \cite{kasiviswanathan2011emerging}, and novel document identification \cite{kasiviswanathan2012online}.

\begin{figure}[t]
\centering
\begin{tikzpicture}
  \begin{axis}[
    thick,
    width=.45\textwidth, height=2cm,
    xmin=-2,xmax=2,ymin=0,ymax=1,
    no markers,
    samples=100,
    axis lines*=left, 
    axis lines*=middle, 
    scale only axis,
    xtick={0},
    xticklabels={},
    ytick={0},
    ] 
  \addplot[red, densely dashed, domain=-2:0]{1*max(-x,0)};
  \addplot[red, densely dashed, domain=0:2]{1*max(x,0)};
  \end{axis}
\end{tikzpicture}
\begin{tikzpicture}
  \begin{axis}[
    thick,
    width=.45\textwidth, height=2cm,
    xmin=-2,xmax=2,ymin=0,ymax=1,
    no markers,
    samples=50,
    axis lines*=left, 
    axis lines*=middle, 
    scale only axis,
    xtick={-1,1},
    xticklabels={$-\kappa$,$+\kappa$},
    ytick={0},
    ] 
\addplot[red,domain=-2:-1,densely dashed]{-x-.5};
\addplot[blue, domain=-1:+1]{.5*x^2};
\addplot[red,domain=+1:+2,densely dashed]{x-.5};
\addplot[blue,mark=*,only marks] coordinates {(-1,.5) (1,.5)};
  \end{axis}
\end{tikzpicture}
\begin{tikzpicture}
  \begin{axis}[
    thick,
    width=.45\textwidth, height=2cm,
    xmin=-2,xmax=2,ymin=0,ymax=1,
    no markers,
    samples=100,
    axis lines*=left, 
    axis lines*=middle, 
    scale only axis,
    xtick={0},
    xticklabels={},
    ytick={0},
    ] 
  \addplot[red, densely dashed, domain=-2:0]{0.6*max(-x,0)};
  \addplot[red, densely dashed, domain=0:2]{1.4*max(x,0)};
  \end{axis}
\end{tikzpicture}
\begin{tikzpicture}
\vspace{-.05 in}
  \begin{axis}[
    thick,
    width=.44\textwidth, height=2cm,
    xmin=-2,xmax=2,ymin=0,ymax=1,
    no markers,
    samples=100,
    axis lines*=left, 
    axis lines*=middle, 
    scale only axis,
    xtick={-.24,.56},
    xticklabels={L, R},
    ytick={0},
    ] 
\addplot[red,domain=-2:-2*0.3*0.4,densely dashed]{2*0.3*abs(x) - 2*0.4*0.3^2};
\addplot[blue,domain=-2*0.3*0.4:2*(1-0.3)*0.4]{0.5*x^2/0.4};
\addplot[red,domain=2*(1-0.3)*0.4:2,densely dashed]{2*(1-0.3)*abs(x) - 2*0.4*(1-0.3)^2};
\addplot[blue,mark=*,only marks] coordinates {(-.24,0.0720) (0.56,0.3920)};
  \end{axis}
\end{tikzpicture}

\begin{tikzpicture}
  \begin{axis}[
    thick,
    width=.45\textwidth, height=2cm,
    xmin=-2,xmax=2,ymin=0,ymax=1,
    no markers,
    samples=50,
    axis lines*=left, 
    axis lines*=middle, 
    scale only axis,
    xtick={-0.5,0.5},
    xticklabels={$-\epsilon$,$+\epsilon$},
    ytick={0},
    ] 
    \addplot[red,domain=-2:-0.5,densely dashed] {-x-0.5};
    \addplot[domain=-0.5:+0.5] {0};
    \addplot[red,domain=+0.5:+2,densely dashed] {x-0.5};
    \addplot[blue,mark=*,only marks] coordinates {(-0.5,0) (0.5,0)};
  \end{axis}
\end{tikzpicture}
\begin{tikzpicture}
  \begin{axis}[
    thick,
    width=.45\textwidth, height=2cm,
    xmin=-2,xmax=2,ymin=0,ymax=1,
    no markers,
    samples=50,
    axis lines*=left, 
    axis lines*=middle, 
    scale only axis,
    xtick={-0.5,0.5},
    xticklabels={$-\epsilon$,$+\epsilon$},
    ytick={0},
    ] 
    \addplot[blue,domain=-2:-0.5] {0.5*(-x-0.5)^2};
    \addplot[domain=-0.5:+0.5] {0};
    \addplot[blue,domain=+0.5:+2] {0.5*(x-0.5)^2};
    \addplot[blue,mark=*,only marks] coordinates {(-0.5,0) (0.5,0)};
  \end{axis}
\end{tikzpicture}
\caption{ Examples of PLQ penalties for dictionary learning, from top to bottom: $\ell_1$-penalty, huber, quantile loss (0.3), quantile huber (0.3)~\cite{AravkinKambadurLozanoLuss2014},
Vapnik, smooth insensitive loss.}
\label{PLQFig}
\end{figure}
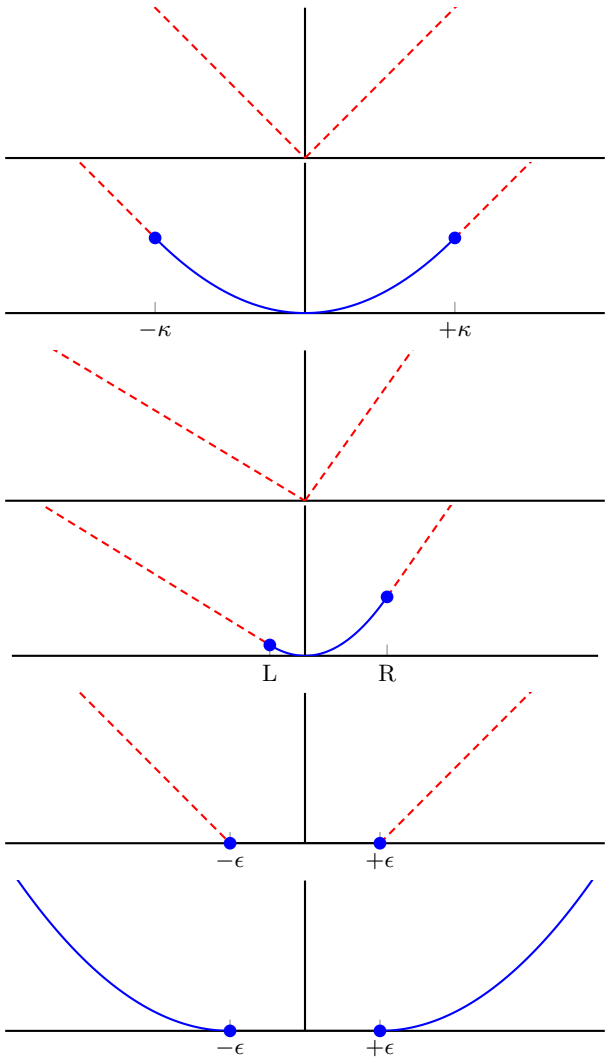

In this paper, we explore and develop a flexible dictionary learning and sparse coding framework,  allowing $\rho_1$ to be a member of a class of functions 
rich enough to address real-world challenges. Such a class should include 
\begin{enumerate}
\item robust penalties, for cases where data may be contaminated by outliers
\item asymmetric penalties, to allow differential treatment of positive and negative elements of the residual vector $r = y-Da$
\item block-assignable penalties, that can act differently on specified subsets of the residual vector
\end{enumerate} 
All of these goals can be achieved by considering the general class of piecewise linear quadratic (PLQ) penalties \cite[Definition 10.20]{RTRW}, 
which comprise convex penalties whose domain can be represented as the union of finitely many polyhedral sets, 
relative to which the penalty can be expressed as a general (convex) quadratic. 
This is a wide class that contains robust penalties such as $\ell_1$, Huber, and Vapnik, 
asymmetric penalties such as quantile \cite{koenker1978regression,takeuchi2006nonparametric}, and quantile Huber~\cite{AravkinKambadurLozanoLuss2014}, 
as well as the classic $\|\cdot\|_2$ penalty, which we refer to as $\ell_2$. 
Some important PLQ penalties are shown in Figure \ref{PLQFig}. 
More details about our proposed framework are available in Section \ref{sec:alg_formulation}. 
Note that learning a dictionary will also be referred to as learning a sparse model in this paper, and without loss of generality, 
we will assume that the regularization $\rho_2$ is the $\ell_1$ measure.

\begin{figure}
\centering
\includegraphics[width = 8cm]{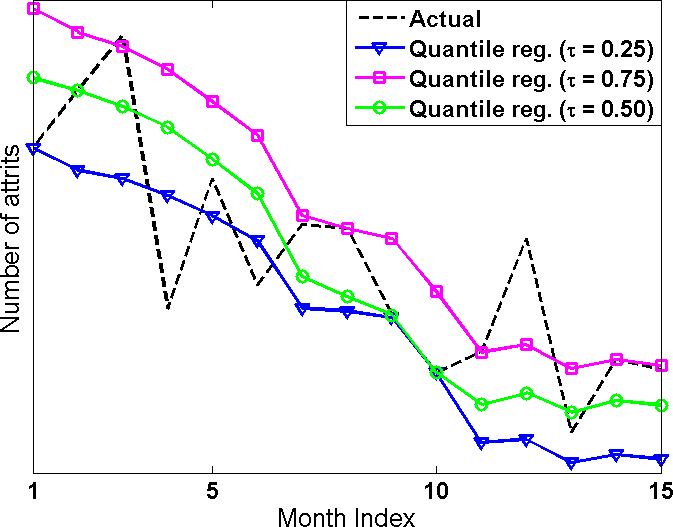}
\caption{The actual attrition and reconstructed attrition of employees in a large organization obtained using quantile regression at the quantiles $\tau = \{0.25, 0.5, 0.75\}$ \cite{Ramamurthy2013GlobalSIP}. The  estimates at $\tau = 0.25$ and $\tau = 0.75$ provides an empirical, non-parametric confidence interval for the median estimate.}
\label{fig:attrn_example}
\end{figure}

\subsection{Going beyond $\ell_2$ loss functions}
\label{sec:beyond_l2_loss}
From a probabilistic viewpoint, if the elements of the residual follow a Gaussian distribution, it is useful to impose a squared error or $\ell_2$ penalty, which is arguably the most widely used loss function. 
However, if we expect outliers in the data, a robust loss function should be imposed; 
and in general, a proper choice of loss function is necessary to estimate the parameters in a noise-robust manner, and in turn lead to improved predictive power for future data. 
Some straightforward examples are: (a) In econometrics where we model the market value of the company as a linear combination of various accounting numbers \cite{barth1998relative} 
and there could be a few bad years for the company due to events like economic depression, 
(b) In image processing where a few pixels are corrupted due to saturation noise from sensors.

Allowing the loss function $\rho_1$ to be asymmetric gives us the ability to penalize positive and negative components of the residual $r = y-Da$ differently. 
Quantile loss is a well-known convex asymmetric function that has been used extensively in regression. 
It is used to understand and predict the response of a process at various quantiles. 
For example, the time-varying attrition of a workforce in a company can be posed as a regression problem over incentive variables, 
and quantile regression allows us to predict the future attrition at various quantiles \cite{Ramamurthy2013GlobalSIP}. 
For planning purposes, management can use best-case and worst case attrition at high and low quantiles, respectively. 
The actual attrition, along with median, high ($0.75$) and low ($0.25$) quantile estimates over a period of time for a particular company are provided in Figure \ref{fig:attrn_example}. 
In addition, predictions at quantiles $0.75$ and $0.25$ can be used to obtain the interquartile range \cite{upton1996understanding}, 
which is a robust measure of statistical dispersion. This provides us with non-parametric, distribution free confidence limit estimates. 
Using our proposed approach, dictionaries can be obtained at various quantiles, which can be then used to obtain such estimates for the predictions.

The case for using a block-assignable penalty for $\rho_1$ may arise when the observation $y$ itself is heterogeneous. 
$\rho_1$ then needs to be constructed to penalize some elements of the residual $r$ differently than the others. 
For example, in image analysis applications, the overall feature $y$ obtained from an image may be a combination of multiple features, each of which require a different loss. 
There could also be situations where a different noise model applies to each component of $y$, and hence different loss functions are needed. 
This is the case in predicting tags for images \cite{}, where $y$ is a combination of image level features (where $\ell_2$ loss can be used) and user tags (where $\ell_1$ loss is required because of the possibility of sparse errors).


\subsection{Contributions}
\label{sec:contrib}

In this paper, we propose a dictionary learning framework for the general class of piecewise linear quadratic (PLQ) penalties, 
and show that the {\it sparse code update problem} for any formulation with PLQ measurement, regularization, and graph structure penalties can be solved 
with a recently developed solver~\cite{JMLR:v14:aravkin13a}. 
For the generalized PLQ approach to the full dictionary learning problem, 
we implement a block-coordinate scheme and prove convergence under the assumption that the measurement PLQ penalty is differentiable. 
Just as in the classic dictionary learning framework, this requires alternating minimization in the sparse codes $A$ and dictionary $D$.
For the latter problem, we use block-coordinate descent to update the columns using an efficient L-BFGS method with Barzilai-Borwein step-length selection.
It is important to note that while this method is general, when $\rho_1$ is the quadratic loss, it converges in two steps (which is almost as efficient 
as a closed-form update rule). 

To enable practitioners to develop and test new kinds of PLQ penalties, we extended the interface of~\cite{ipSolve:2012} to allow 
(1) different PLQ penalties for different blocks of a residual vector, and 
(2) automatic Moreau-Yosida smoothing of arbitrary PLQ penalties.
The latter feature ensures convergence of block coordinate descent, if applied to any (potentially non-smooth) PLQ formulation.  
These extensions are communicated as theoretical lemmas related to conjugate representation calculus. 

To illustrate the utility of the proposed approach, we apply the algorithm in three different real-world scenarios and provide experimental evaluations. 
The first scenario concerns the robust modeling of images corrupted by sparse noise. 
In this case, we train a dictionary for sparse coding of the patches, taking $\rho_1$ to be the robust Huber penalty. 
The dictionary is then used to reconstruct the patches, and this allows much better noise rejection compared to recovery obtained with $\rho_1$ set as the $\ell_2$ penalty. 
In the second case, we consider the problem of refining human annotated tags in an image data set. 
Since tags of similar images will be similar, we perform a joint sparse coding of features and tags using the mixed PLQ penalty, 
with $\ell_2$ penalty for features and Huber penalty for tags. 
The mixed penalty provides more robust estimates compared to using $\ell_2$ for both features and tags, at varying levels of impulse noise.
In the third application, we evaluate the  performance of subspace clustering using $\ell_1$ graphs \cite{l1graph} in various data sets at multiple quantiles. 
The performance of different quantiles around the median can be used to evaluate empirical confidence bounds on the median accuracy. 

\section{Algorithmic Formulation}
\label{sec:alg_formulation}

We begin by formulating a generalized batch dictionary learning problem:
\begin{equation}
\label{eq:full}
\begin{aligned}
\min_{A, D}\quad &\rho_1(Y - DA) + \rho_2(A) + \rho_3(A)\\ 
& \text{subject to } A \in \mathcal{A}, D \in \mathcal{D}.
\end{aligned}
\end{equation}
where $Y = [y_1 y_2 \ldots y_T]$ is the observation matrix, $A = [a_1 a_2 \ldots a_T]$ is the corresponding sparse code matrix, 
$\rho_1$ is the misfit loss function 
$\rho_2$ is the sparsity regularization, and $\rho_3$ encodes other prior information about the codes (for example, graph structure). 
The constraints $A \in \mathcal{A}$ and $D \in \mathcal{D}$ allow us to encode other prior information about the codes and the dictionary; 
for example, the columns of the dictionary may be normalized, while all codes may be non-negative.

This problem is nonconvex, and is typically solved in a block-coordinate descent fashion, where the dictionary $D$ and codes $A$ are updated in turn, with the other held fixed. Note that for fixed $D$, the problem of updating $A$ fully decouples when $\rho_1$ is the Frobenius norm, and every column of $A$ may be updated in parallel. 
This generalizes perfectly for any loss function $\rho_1$ that can be written as a sum of penalties across columns; 
we assume our loss functions have this property. 
We refer to this problem as the {\it code update problem}. For fixed $A$, the problem of updating $D$ requires consideration of $\rho_1$ only, 
together with the constraint $D \in \mathcal{D}$. We refer to this as the {\it dictionary update problem}.

In this paper, we propose a modeling framework and optimization scheme 
that is general enough to handle all of these requirements, as well as simple constraints
on $A$ and $D$. Specifically, we allow $\rho_1$ to come from the class of 
piecewise linear quadratic (PLQ) penalties, or a mixture of several PLQ penalties. Recently,~\cite{JMLR:v14:aravkin13a} showed that a broad subclass of
these penalties can be given a natural statistical interpretation, and used their conjugate representation to devise a generic interior point method for their solution. This method also efficiently incorporates simple constraints $A \in \mathcal{A}$. We use this method to solve the code-update problem.  

In the next section, we review the general class of PLQ penalties, and characterize
some properties of this class that make it particularly useful for modeling 
specific applications. We then specify the representations of penalties we present in our experimental section. We then discuss our method of solving the dictionary update problem, and consider convergence of the entire scheme.

\subsection{Piecewise Linear-Quadratic penalties}

We briefly review the class of quadratic support (QS) functions, 
referring the reader to~\cite{JMLR:v14:aravkin13a} for a full exposition. Every penalty in 
this class can be written as a convex conjugate to a quadratic function on an arbitrary set: 
\begin{definition}
A QS function is any function $\rho(U, M, b, B; \cdot)$ 
mapping from $\mB{R}^n$ to $\mathbb{\overline R} = \mB{R} \cup \infty$ 
having representation
\begin{equation}\label{PLQpenalty}
\rho(U, M, b, B; y) 
=
\sup_{u \in U}
\left\{ \langle u,b + By \rangle - \half\langle u, Mu
\rangle \right\} \;,
\end{equation}
where $U \subset \mB{R}^m$ is a nonempty convex set, 
$M\in \Snp$ the set of real symmetric positive semidefinite matrices,
and $b + By$ is an injective affine transformation in $y$, with $B\in\mB{R}^{m\times n}$, 
so, in particular, $m \leq n$ and $\R{null}(B) = \{0\}$. 
\end{definition}
If the set $U$ is taken to be polyhedral, i.e. having the representation 
\[
U := \{u|Cu \leq c\},
\]
then the associated QS function becomes piecewise linear quadratic, and is written
\(\rho(C, c, M, b, B; y)\). 

The ability to represent PLQ penalties through the set of structures $(A, a, M, b, B)$
gives rise to a representation calculus, where addition, affine composition, and 
other manipulations can be done using the underlying structures. We highlight 
three results that are particularly useful in encoding variants of the code update problem. 

\begin{lemma}[Addition]
\label{lem:addition}
Let $\rho_1(y)$ and $\rho_2(y)$ be two PLQ penalties specified by
$C_i, c_i, M_i, b_i, B_i$, for $i = 1, 2$. Then the sum 
$\rho(y) = \rho_1(y) + \rho_2(y)$
is also a PLQ penalty, with 
\[
C = \begin{bmatrix} C_1 & 0 \\ 0 & C_2\end{bmatrix},\; a = \begin{bmatrix} a_1 \\ a_2 \end{bmatrix},\;
M = \begin{bmatrix} M_1 & 0 \\ 0 & M_2\end{bmatrix},\;
b = \begin{bmatrix} b_1 \\ b_2\end{bmatrix},\;
B = \begin{bmatrix} B_1 \\ B_2 \end{bmatrix}\;. 
\]
\end{lemma}

This result is used both to combine measurement and regularization terms 
into a single representation. 

\begin{lemma}[Affine composition]
\label{lem:composition}
PLQ penalties are closed under affine composition, with 
\[
\rho(C, c, M, b, B; Dy+d) = \rho(C, c, M, b+Bd, BD; y).
\]
\end{lemma}

This result allows automatic composition of simple building blocks with linear maps. 

\begin{lemma}[Product action]
\label{lem:product}
A PLQ $\rho(y) = \rho_1(y_{1}) + \rho_2(y_{2})$, where $y_1$ and $y_2$ are sub-blocks of the 
vector $y$, is easily written in terms of addition and affine composition; namely  
\[
\rho(y) = \rho_1(M_1 y) + \rho_2(M_2 y),
\]
where $M_1y = y_1$ and $M_2y = y_2$. 
\end{lemma}

This last result makes it easy to use different penalties on different 
variable or residual subsets, which is important in some of our applications. 

\subsection{Optimizing PLQ penalties with polyhedral constraints }

Lemmas~\ref{lem:addition},~\ref{lem:composition} and~\ref{lem:product} show that inference problems 
involving sums, affine compositions, and coordinate-wise different PLQ penalties can at the end of the day 
be written down as a minimization problem in the primal variable $y$ and the {\it conjugate} variable $u$. 
To this optimization problem we now add polyhedral inequality constraints $Ay \leq a$, obtaining the 
most general model problem:
\[
\min_{y, u} \rho(C, c, M, -b, -B; y) \quad \text{s.t. } Ay \leq a.  
\]

Note that a simple {\it evaluation} of our function at a candidate point $y$
requires partial minimization with respect to the conjugate variable $u$. 
It may therefore seem that we have made the problem more complicated; 
however, keep in mind that (a) if we choose component PLQ penalties 
from a wide set of common candidates, we have alternative representations
for $\rho(y)$ at our disposal, and (b) the conjugate representation has been
introduced for the purpose of obtaining a minimum in $y$. 

Through the conjugate representation, we are able 
to write the Karush-Kuhn-Tucker system of optimality conditions
for the entire class of interest.
KKT systems are often used to characterize optimality of optimization programs 
and design algorithms; the advantage of the conjugate representation is that 
we have a uniform approach to characterizing a wide variety of nonsmooth optimization
programs.  The details are encoded within the representation, which is formed 
automatically from individual components using the calculus 
we have described in the previous section.
The KKT system is
\begin{equation}
\label{fullKKT}
\begin{aligned}
0 &= B^\R{T}u +Aw\\
0 &= b+By - Mu - Aq \\
0 &= C^{T}u + s - c\\
0 &= A^Ty + r - a\\
0 &= q_is_i \;, \  i=1,\dots,\ell\;, \; q, s \geq 0\;,\\
0 &= w_ir_i \;, \  i=1,\dots,\ell\;, \; w, r \geq 0\;,
\end{aligned}
\end{equation}
where $s, r$ are nonnegative slack variables that turn inequality constraints $C^Tu \leq c$ and 
$A^Ty \leq a$ into equations, while $q, w$ are the dual variables corresponding to the 
resulting equality constraints. With~\eqref{fullKKT} in hand, the problem can be 
solved by relaxing the complementarity slackness conditions (last two equations), 
and using a damped Newton's method to directly optimize the relaxed system. 
Full convergence theory for problem without inequality constraints is presented by~\cite{JMLR:v14:aravkin13a},
and~\cite{AravkinBurkePillonetto2013b} shows how constraints can be included. 

In the context of dictionary learning, any PLQ penalty can be used for the sparse 
code update problem. Since we directly 
solve the KKT system~\ref{fullKKT} using the method of~\cite{JMLR:v14:aravkin13a}, 
we always have direct access to an optimality certificate; namely the KKT system itself. 
However, these guarantees only hold for the (convex) code update problem, and in
the next section, we discuss the overall (nonconvex) approach to dictionary learning, and 
focus on the dictionary update problem.

\subsection{Block coordinate descent}

We now consider the full nonconvex problem~\eqref{eq:full}. A natural approach
is to alternate between updating spare codes $A$ and the dictionary $D$, 
which is an instance of block coordinate descent.  
When the penalties $\rho_1$, $\rho_2$ and $\rho_3$ are smooth, 
standard convergence results for block coordinate descent can be obtained with e.g.~\cite[Proposition 2.7.1]{Bert}).
However, in most sparse dictionary learning, $\rho_2$ is taken to be non-smooth, usually the $\ell_1$ norm.
In addition, we are interested in a general theory that applies to the entire PLQ class. 

Block-coordinate descent for a class of problems general enough to accommodate our framework 
is studied in~\cite{Tseng2001}. The main theorem of~\cite{Tseng2001} still depends on a smoothness condition 
for $\rho_1$, and, unfortunately, as~\cite{Tseng2001} points out, this condition is in some sense sharp, because
block coordinate descent can fail to converge if only convexity in each block is required. 

We now present the main theorem for batch dictionary learning in the PLQ setting. 
\begin{theorem}
\label{thm:diffble}
Suppose that $\rho_1$ in~\eqref{eq:full} is differentiable, $\rho_2$ and $\rho_3$ are convex, 
and the sets $\mathcal{D}$ and $\mathcal{A}$ are convex. Then block coordinate descent 
(alternating minimization in $A$ and $D$) converges to a stationary point of~\eqref{eq:full}.
\end{theorem}

\proof{By assumption, $\rho_1$ is differentiable 
on its effective domain; furthermore, the entire objective is convex in $A$. By~\cite[Lemma 3.1 and Theorem 4.1(b)]{Tseng2001}, 
every cluster point of the sequence generated from block-coordinate descent is a stationary point of~\eqref{eq:full}.}

From the application perspective, we claim that the requirement that $\rho_1$ be smooth 
is not particularly limiting. To understand why, recall that the behavior of a penalty at the origin 
has the strongest influence on {\it sparsity} properties; in particular, this is why the $\ell_1$ 
penalty is a key choice for $\rho_2$. In contrast,  $\rho_1$
acts on {\it residuals}; the implication is that choosing $\rho_1$ that are non smooth (say at the origin)
means we will fit a lot of data points {\it exactly}. While this is potentially useful in some applications,
it is not in others;~\cite{AravkinKambadurLozanoLuss2014} recently demonstrated that 
a smoothed version of the quantile penalty called the {\it quantile huber} can outperform
the standard quantile penalty in the sparse regression setting. 

The next natural question is, suppose we are given a PLQ penalty candidate for $\rho_1$ 
which is not smooth; is there a disciplined procedure we can use to smooth it, and remain in the class? 
Amazingly, it turns out that any PLQ penalty can be easily smoothed using a Moreau envelope, and moreover,
application of this technique can be represented using the same calculus we relied on in Lemmas~\ref{lem:addition}, 
\ref{lem:composition}, and \ref{lem:product}.

Recall that the Moreau envelope of a convex function $g$ can be defined as follows:
\begin{equation}
\label{eq:MY}
e_{\gamma}g(y) = \min_x \frac{1}{2\gamma}\|x-y\|^2 + g(x). 
\end{equation}
From the definition, it is immediately clear that $g_\rho(y)$ is well defined, and is always lower 
than $g$, because $y$ is feasible in~\eqref{eq:MY}. Finally, it can be easily shown~\cite{RTRW}
that $g_\rho(y)$ is differentiable, with gradient given by 
\[
\nabla g_\gamma(y) = \frac{1}{\gamma}(y - \bar x),
\]
where $\bar x$ is the unique minimizer of~\eqref{eq:MY}, and $\gamma$ is a smoothing parameter. The Moreau-Yosida envelope function is also know as 
the prox operator, and  plays a major role in optimization formulations for many signal processing applications, c.f.~\cite{combettes}. 
As $\gamma \downarrow 0$, the envelope function converges to $g$ in an epigraphical sense. 

The salient feature for us is that PLQ penalties are closed under Moreau-Yosida smoothing, 
and the envelope function is precisely captured by our PLQ representation, as shown in~\cite[Proposition 4.11]{BurkeH13}, 
reproduced here for convenience in our notation:

\begin{theorem}
Let $\rho(c, C, M, b, B; y)$ be any PLQ penalty. Then the envelope function $e_\gamma\rho(y)$ is
also a PLQ penalty with representation $\rho(c, C, \overline{M}, b, B; y)$, where 
\[
\overline{M} = M + \gamma BB^T. 
\]
\end{theorem}

This theorem allows us not only to smooth any member of the PLQ class by a particular amount $\gamma$, 
but to obtain the representation of the resulting function in closed form. The power of this idea is shown by the following
corollary: 
\begin{corollary}
The envelope of the 1-norm,  $e_\gamma \ell_1$, is the huber function with threshold $\mu$. 
\end{corollary}

In particular, the `smoothing' of the $\ell_1$ to get the huber is an idea that easily generalizes to the entire class, 
and is captured by the conjugate representation calculus. 

\begin{figure*}[t]
\centering
\begin{subfigure}[b]{0.32\textwidth}    
 	\includegraphics[width=5.5cm]{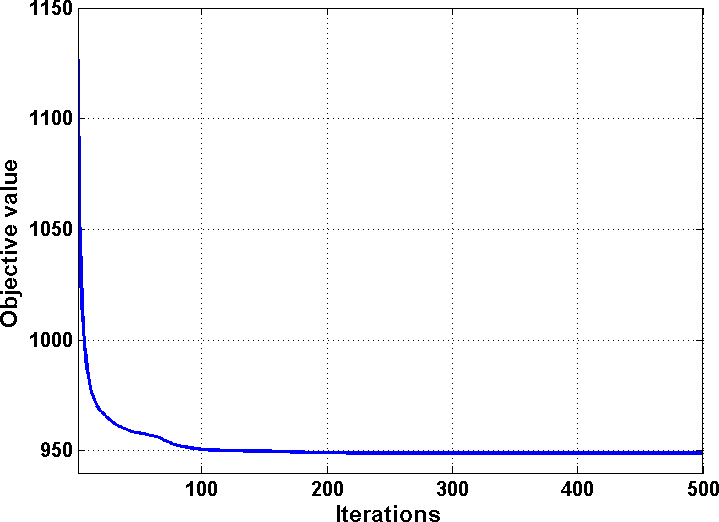}
 	\caption{$\ell_2$ misfit loss}           
\end{subfigure}%
\quad
\begin{subfigure}[b]{0.32\textwidth}
 	\includegraphics[width=5.5cm]{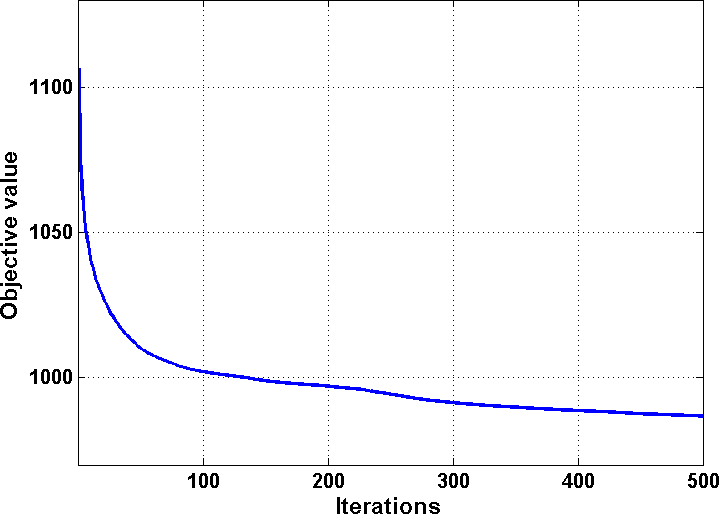}     
 	\caption{Huber misfit loss}         
\end{subfigure}%
\quad
\begin{subfigure}[b]{0.32\textwidth}
 	\includegraphics[width=5.5cm]{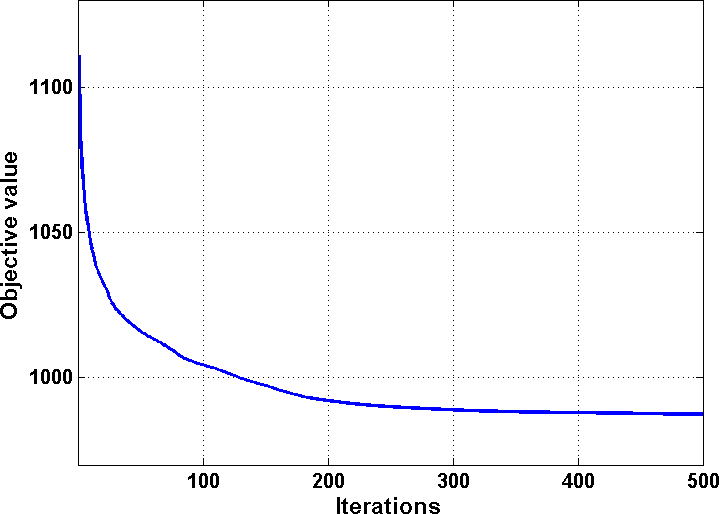}      
 	\caption{Quantile Huber misfit loss ($\tau = 0.25$)}       
\end{subfigure}%
\caption{Empirical convergence of the block-coordinate descent scheme for the proposed batch dictionary learning problem for various misfit losses.}
\label{fig:conv_plots}
\end{figure*}

\subsection{Dictionary update problem}

In the previous section, we simply claimed to solve the dictionary update problem
\[
\min_D \rho_1(Y - DA)
\]
for a fixed set of sparse codes $A$. We now explain how we solve this problem, and prove the 
convergence of our scheme. In the least squares case, it is straightforward to implement a block-coordinate
optimization scheme on the columns of $D$, obtaining closed-form updates as we loop over the columns. 

In the general case, up pose that we wish to update the $j$-th column of $D_{j}$. 
Letting $a_j$ denote the $j$th {\it row} of $A$, $d_j$ denote the $j$th column of $D$, and 
$D_{/j}$ to denote the dictionary with the $j$th column deleted, we it is easy to see that  
\[
DA - D_{/ j}A  = d_j a_j^T. 
\]
For penalties $\rho$ which decompose over the columns of the residual $Y-DA$, 
the optimization formulation to $d_j$ is given by 
\begin{equation}
\label{eq:column}
\bar d_j = \min_d \rho_1(Y_j - d a_j^T). 
\end{equation}
with $Y_j = Y - D_{/j}A$. For the least-squares case, this update problem has a closed form solution;
and in the general case, the structure of the problem is very simple: the $k$th entry of $d_{j,k}$
is determined by solving a scalar optimization problem
\[
d_{j,k} = \argmin_{d_k} \rho_1(Y_j(k, \cdot) - d_k a_j).
\]
Since this is a 1-dimensional optimization problem, the Barzilai-Borwein~\cite{BARZILAI} line search method 
is equivalent to Newton's method in the quadratic case (after 2 steps). 
Motivated by this, we use L-BFGS with Barzilai-Borwein line search to solve~\eqref{eq:column}.
For quadratic $\rho_2$, this method converges in two iterations per column, as expected, and for general smooth $\rho_2$,
such as the Huber, it is also rapidly convergent. Since Theorem~\ref{thm:diffble} requires $\rho_1$ to be smooth, 
block-column coordinate descent converges by~\cite[Proposition 2.7.1]{Bert}.

The empirical convergence of the overall block-coordinate descent scheme for the proposed batch dictionary learning problem is shown in Figure \ref{fig:conv_plots}, for a real-world data set. Clearly, the $\ell_2$ loss has the fastest convergence when compared to Huber and quantile Huber ($\tau = 0.25$), since it is the most well-behaved among all the three.

\begin{figure*}[t]
\centering
\begin{subfigure}[b]{0.18\textwidth}
	\center{PSNR = 25.32 dB}    
 	\includegraphics[width=3.4cm]{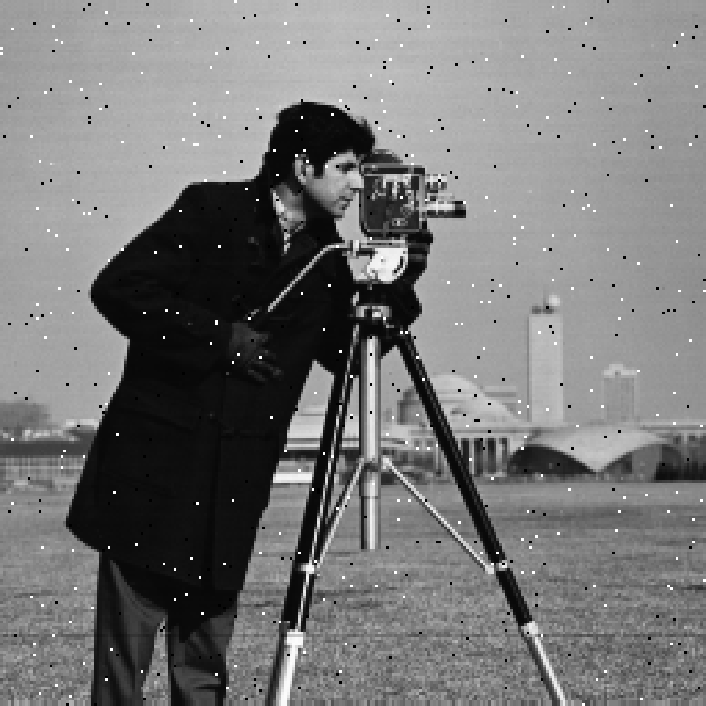}           
\end{subfigure}%
\quad
\begin{subfigure}[b]{0.18\textwidth}
 \center{PSNR = 20.46 dB} 
 	\includegraphics[width=3.4cm]{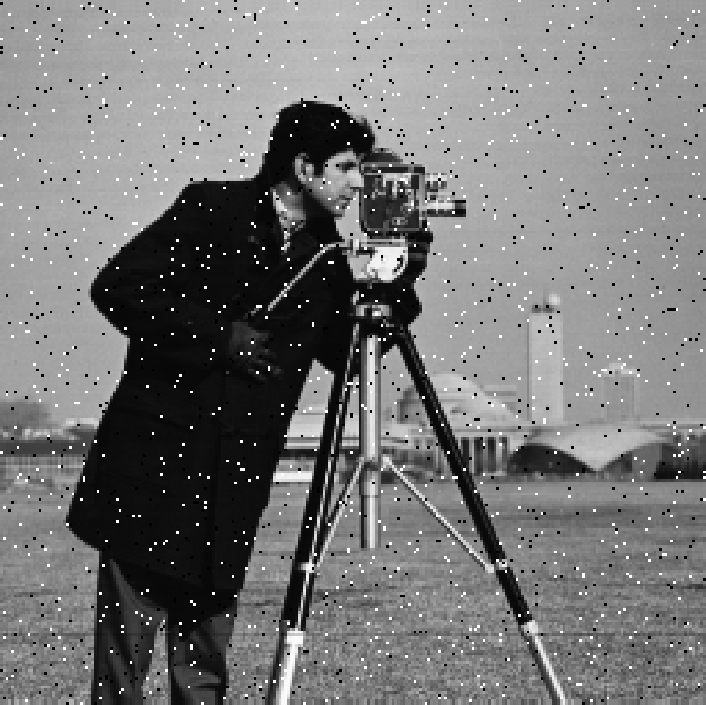}             
\end{subfigure}%
\quad
\begin{subfigure}[b]{0.18\textwidth}
 \center{PSNR = 18.25 dB}      
 	\includegraphics[width=3.4cm]{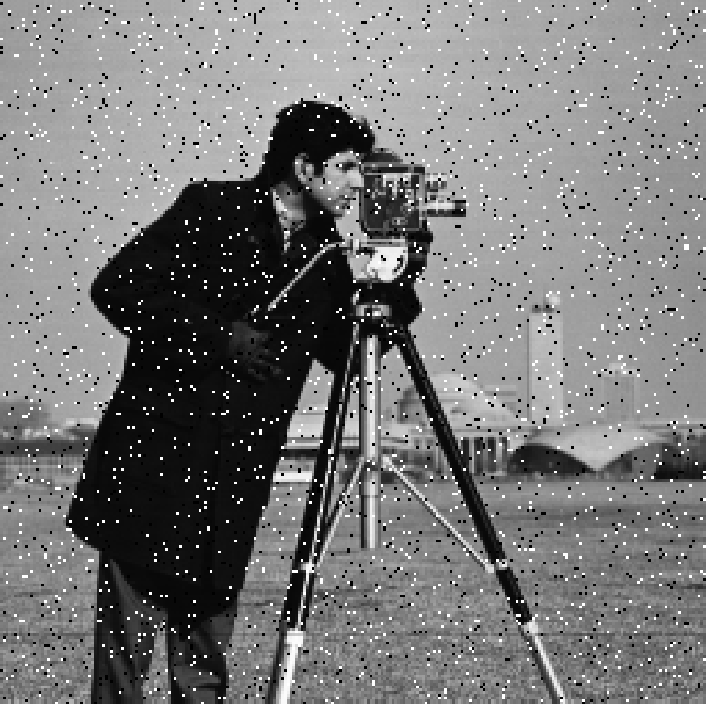}            
\end{subfigure}%
\quad
\begin{subfigure}[b]{0.18\textwidth}
	\center{PSNR = 15.18 dB}  
 	\includegraphics[width=3.4cm]{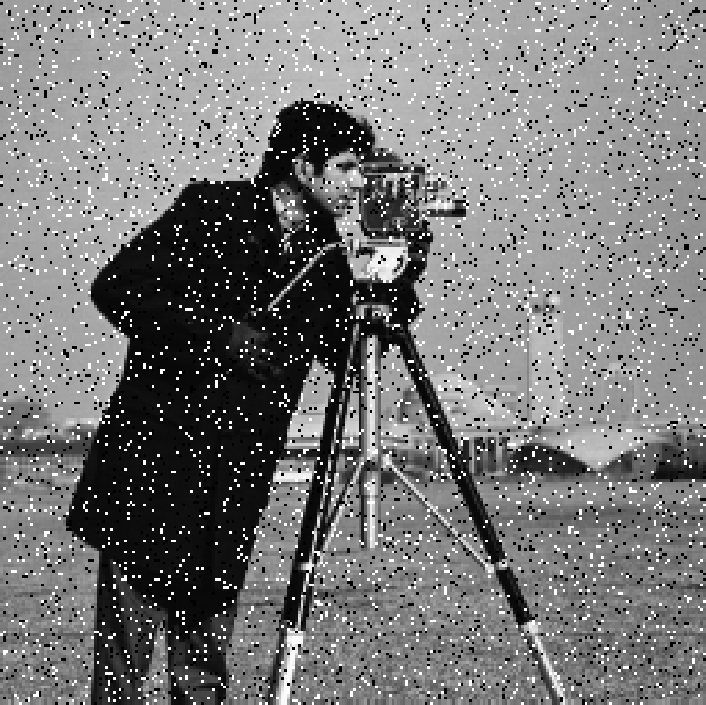}             
\end{subfigure}%
\quad
\begin{subfigure}[b]{0.18\textwidth}
	\center{PSNR = 13.35 dB}  
 	\includegraphics[width=3.4cm]{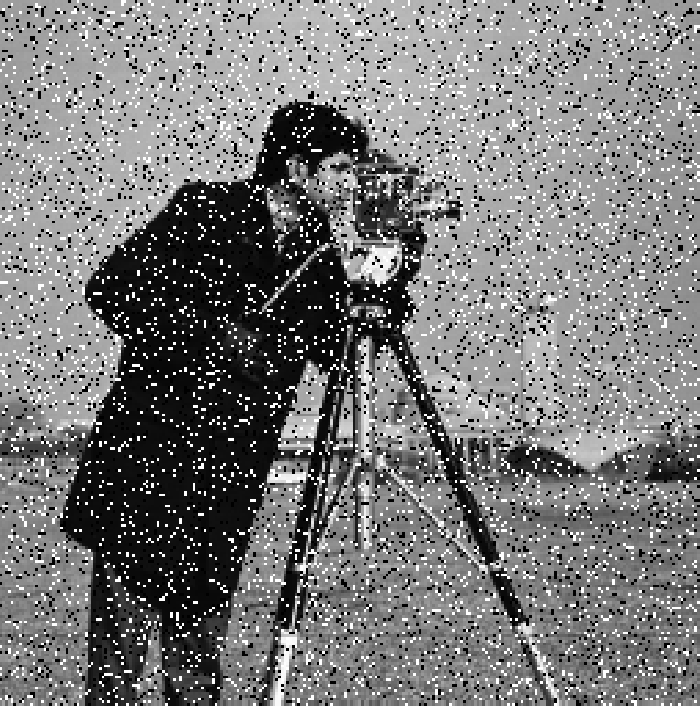}           
\end{subfigure}%
\\
\vspace{0.2in}
\begin{subfigure}[b]{0.18\textwidth}
	     \center{PSNR = 28.58 dB}   
 	\includegraphics[width=3.4cm]{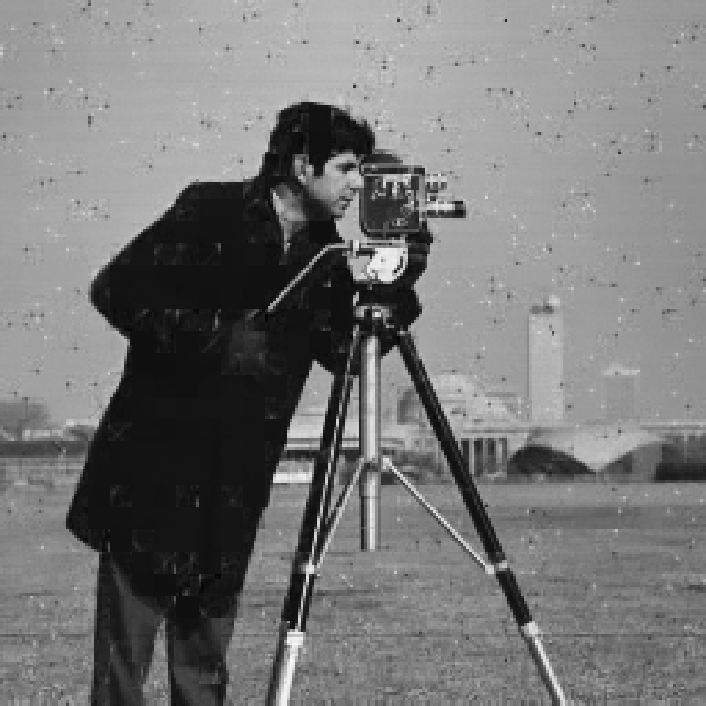}            
\end{subfigure}%
\quad
\begin{subfigure}[b]{0.18\textwidth}
	     \center{PSNR = 25.96 dB}   
 	\includegraphics[width=3.4cm]{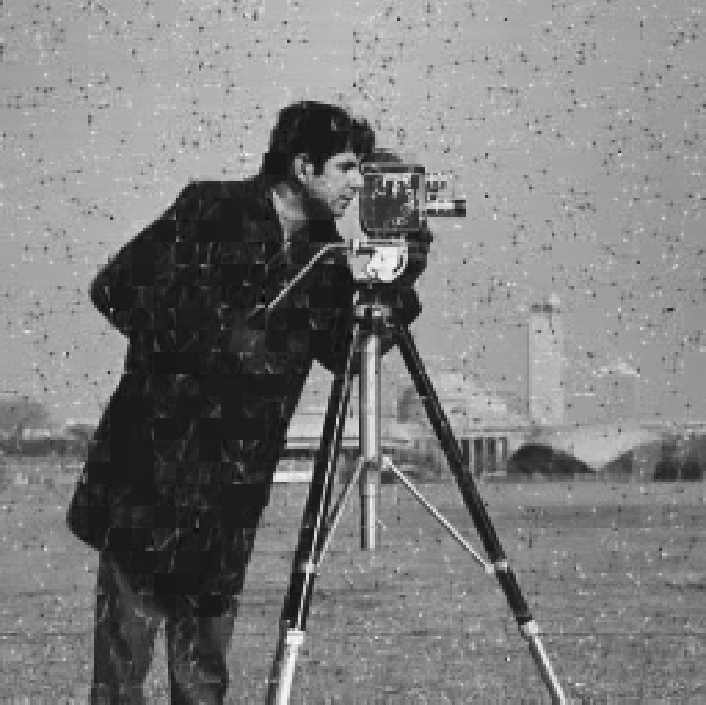}            
\end{subfigure}%
\quad
\begin{subfigure}[b]{0.18\textwidth}
	     \center{PSNR = 24.29dB}   
 	\includegraphics[width=3.4cm]{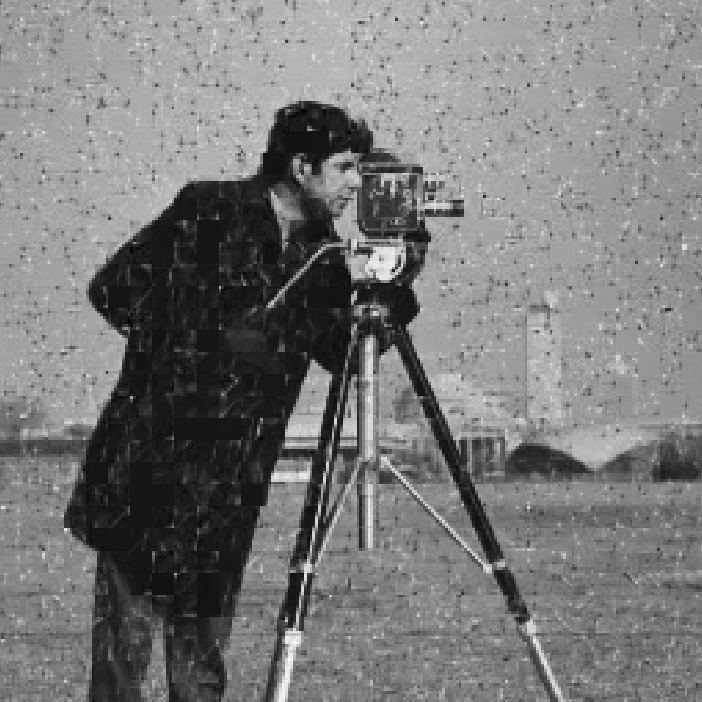}           
\end{subfigure}%
\quad
\begin{subfigure}[b]{0.18\textwidth}
     \center{PSNR = 21.54 dB}   
 	\includegraphics[width=3.4cm]{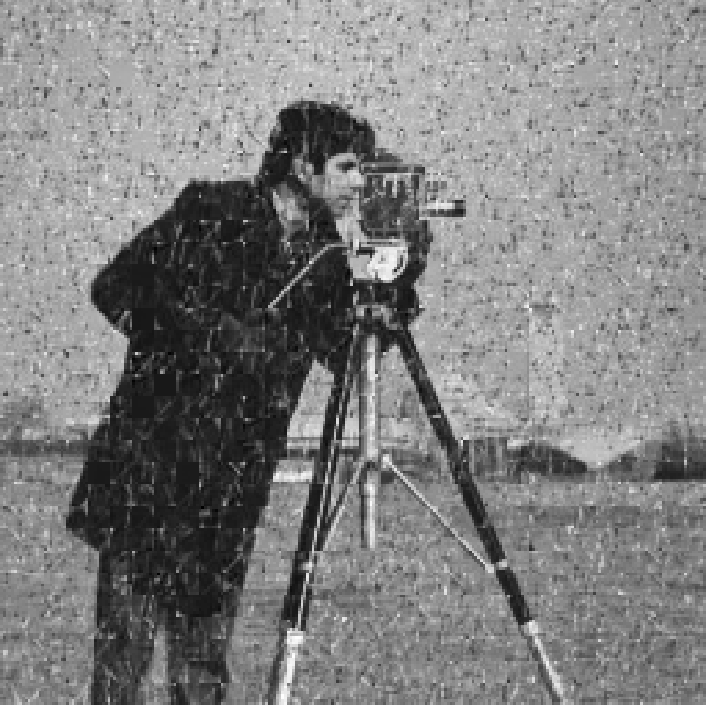}             
\end{subfigure}%
\quad
\begin{subfigure}[b]{0.18\textwidth}
     \center{PSNR = 19.71 dB}    
 	\includegraphics[width=3.4cm]{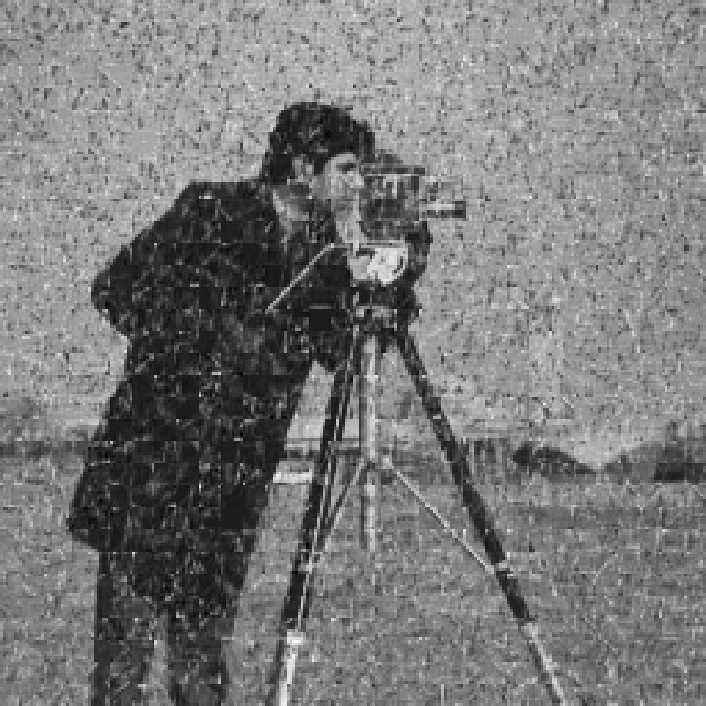}           
\end{subfigure}%
\\
\vspace{0.2in}
\begin{subfigure}[b]{0.18\textwidth}
	\center{PSNR = 35.08 dB}    
 	\includegraphics[width=3.4cm]{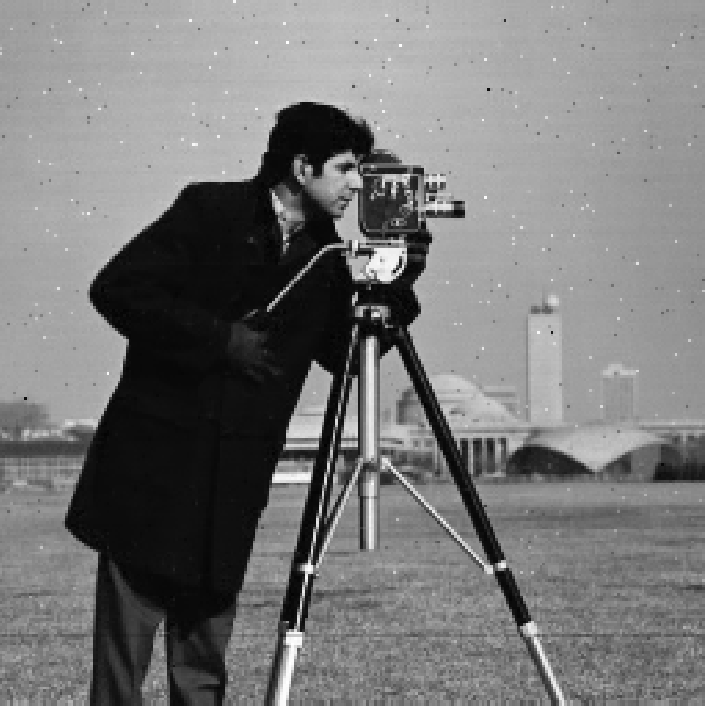}            
\end{subfigure}%
\quad
\begin{subfigure}[b]{0.18\textwidth}
\center{PSNR = 31.42 dB}   
 	\includegraphics[width=3.4cm]{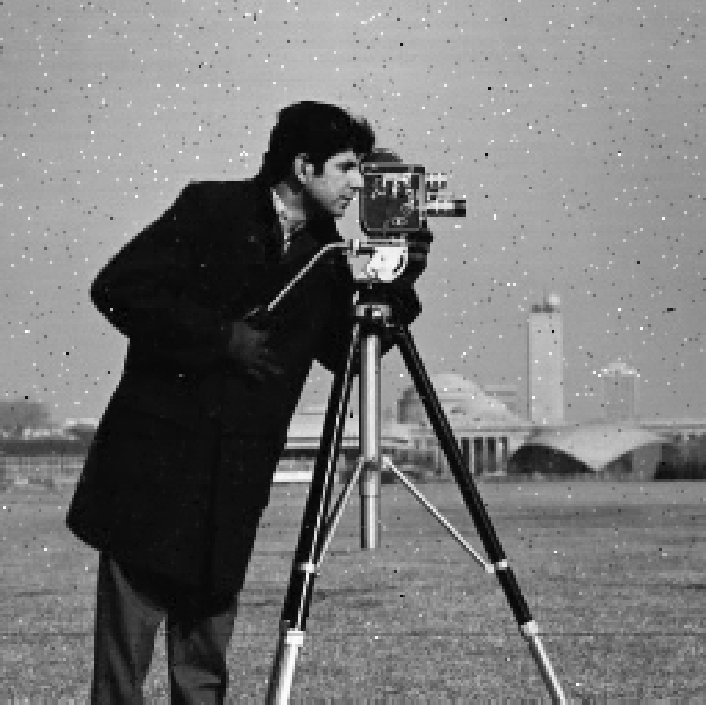}          
\end{subfigure}%
\quad
\begin{subfigure}[b]{0.18\textwidth}
\center{PSNR = 29.48 dB}   
 	\includegraphics[width=3.4cm]{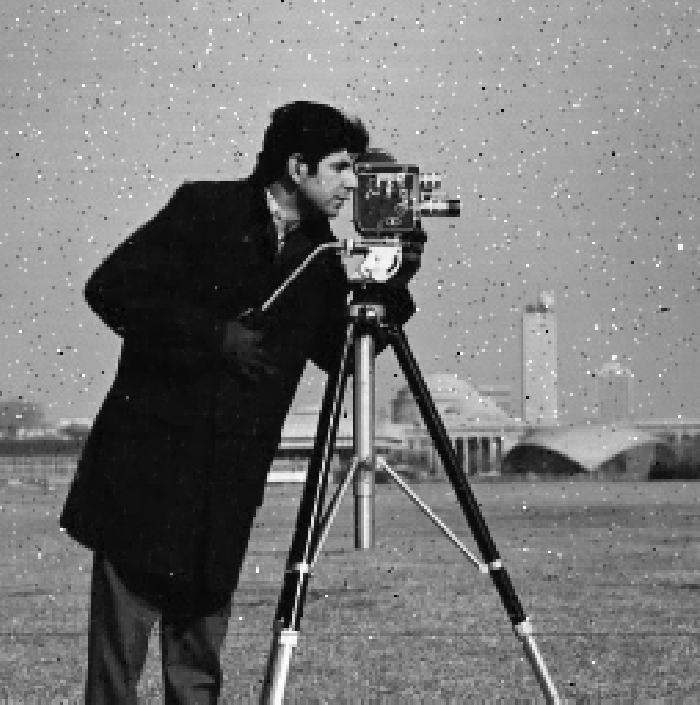}          
\end{subfigure}%
\quad
\begin{subfigure}[b]{0.18\textwidth}
\center{PSNR = 25.83 dB}   
 	\includegraphics[width=3.4cm]{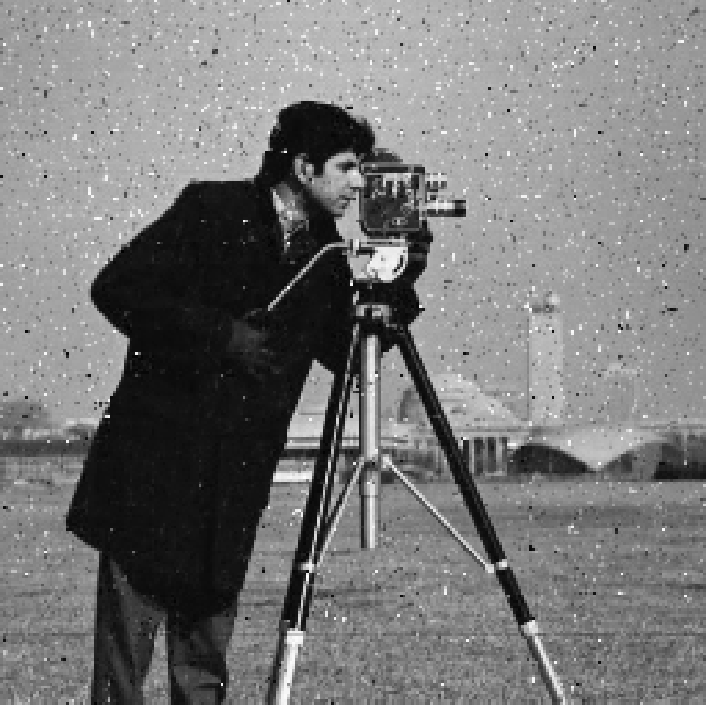}      
\end{subfigure}%
\quad
\begin{subfigure}[b]{0.18\textwidth}
\center{PSNR = 22.94 dB}   
 	\includegraphics[width=3.4cm]{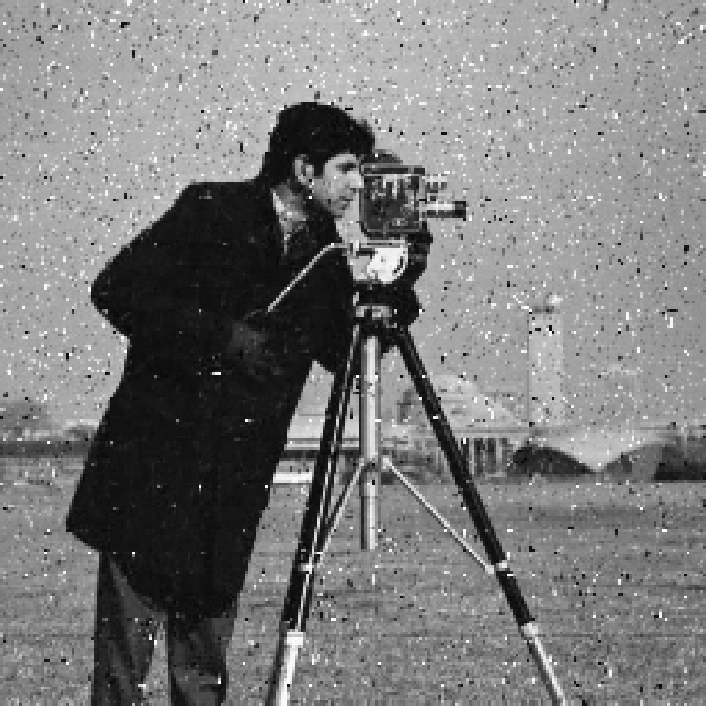}           
\end{subfigure}%
\caption{Robust Image Modeling - Row $1$ shows images corrupted by increasing levels of salt and pepper noise. 
Rows $2$ and $3$ show the images recovered using sparse models learned with the $\ell_2$ and Huber penalties, respectively. 
In each case, the corresponding PSNR (dB) value is also reported. 
The robustness achieved by considering a more appropriate loss function when the corruption is non-Gaussian is clearly evident.}
\label{fig:imgmodel}
\end{figure*}

\section{Experiments}
\label{sec:results}
Added flexibility in the choice of loss functions can make sparse models more effective in several applications. 
In this section, we present three different case studies that demonstrate the importance of adopting a general framework.
Though more sophisticated applications of sparse models can be considered, 
our emphasis is on illustrating the flexibility and robustness of the proposed framework in comparison to conventional sparse modeling approaches. We conclude this section by briefly discussing possible extensions to this work, and relevant applications that can benefit from the general framework developed in this paper.

\subsection{Scenario 1: Robust Image Modeling}
The statistics of natural images motivates the use of sparse models to describe them \cite{Field1987}, 
and makes it possible to recover them from different forms of corruption. 
For simplicity, we consider corruption by an additive noise, which can happen during sensing or transmission. 
In such scenarios, a generalizable model should ignore the underlying noise, and describe only the relevant patterns in the image. 
Such a robust model can then be used to denoise the image or improve its quality. 
When the noise is Gaussian, the traditional sparse modeling framework, which uses an $\ell_2$ loss function, can be very effective in discovering patterns that are masked by noise. 
However, when the noise model is non-Gaussian, the sparse model learned using this procedure will no longer be robust.

Consider the case where an image is corrupted by salt-and-pepper noise, which manifests as randomly occurring white and black pixels in the image. 
A typical noise reduction strategy for this kind of noise is to apply median filtering. 
Therefore, we propose to use the Huber penalty as the loss function, since it can learn median patterns in the dictionary, thereby resulting in a robust model.  
However, using an $\ell_2$ loss function will infer noisy patterns, since it tends to spread the noise through the pattern, and hence the denoising performance 
using the dictionary will also be poor.

\begin{figure*}[t]
\centering
\begin{subfigure}[b]{0.32\textwidth}    
 	\includegraphics[width=5.5cm]{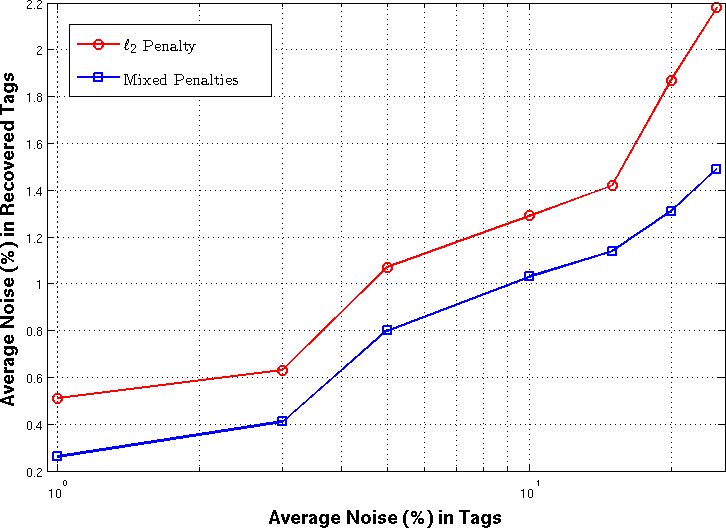}
 	\caption{$0\%$ Training Noise}           
\end{subfigure}%
\quad
\begin{subfigure}[b]{0.32\textwidth}
 	\includegraphics[width=5.5cm]{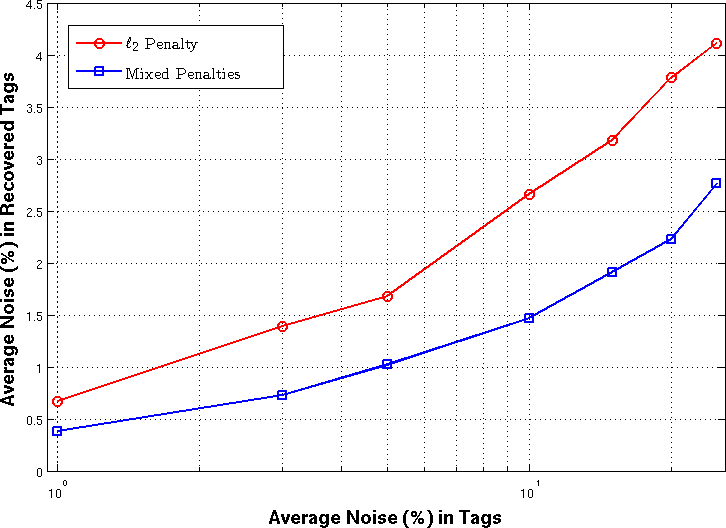}     
 	\caption{$5\%$ Training Noise}         
\end{subfigure}%
\quad
\begin{subfigure}[b]{0.32\textwidth}
 	\includegraphics[width=5.5cm]{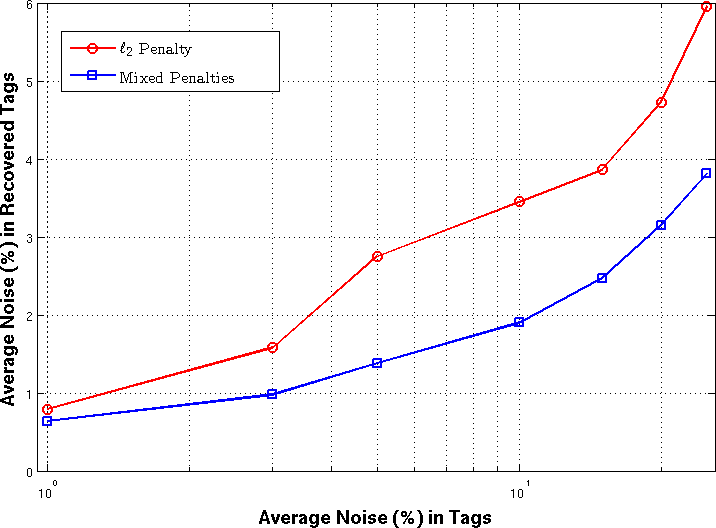}      
 	\caption{$10\%$ Training Noise}       
\end{subfigure}%
\caption{Tag Refinement using $\ell_2$ (red) and mixed $\ell_2-$Huber penalties (blue). 
Using appropriate robust penalties for the tags result in improved recovery performance at all noise levels for training data.}
\label{fig:tags}
\end{figure*}

Given an image $I$, we extract non-overlapping patches of size $8 \times 8$, vectorize these patches, and stack them to form a matrix denoted $X$. 
Adding salt-and-pepper noise at a specific percentage is equivalent to randomly replacing the percentage of pixels with a black or a white pixel. 
In this experiment, we vary the noise level from $1\%$ to $15\%$ . 
We learn dictionaries using different penalties, and compare the reconstruction obtained using the learned sparse model with the original clean image. 
When the model is robust, we expect that the impulse noise will not be a part of the dictionary elements, and hence the reconstruction will be of high quality. 
Note that we do not perform any explicit denoising, and only evaluate the quality of the reconstruction from the model. 
We measure the peak-signal-to-noise ratio (PSNR) for the noisy image, and the images recovered using sparse models learned with $\ell_2$ and Huber penalties. 
Figure \ref{fig:imgmodel} shows the results obtained for increasing levels of impulse noise, 
and the robustness of the Huber penalty is clearly evident from the higher PSNR values, as well as from the improved visual quality.

\subsection{Scenario 2: Refining Tags for Image Retrieval}
Textual descriptors, or {\it tags}, are useful meta-data for images in retrieval applications. 
In large scale retrieval systems, it is typical to present a textual query to retrieve semantically relevant images. 
Since a single semantic concept can manifest into a wide range of visual representations, 
it is often difficult to mine a database using just visual features or tags. 
Furthermore, human annotation can be very subjective and error-prone. 
The goal of automatic image annotation is to predict new tags, and possibly refine existing noisy tags, based on information from visually similar images. 
In this experiment, we will consider the problem of refining the noisy tags of a novel image using a set of training images. 
For each image, a tag vector is typically a binary vector that indicates the relevance of each semantic topic from a pre-defined vocabulary. 
Due to human errors, or the limitation of prediction systems, some unrelated concepts could be included in the image description, or important topics could be left out. 
Sparse or low-rank models learned using both visual features and the noisy tags can be very effective in refining the semantic descriptors \cite{wang2009multi,zhang2010automatic,zhu2010image}.

Given a set of training images, we use the \textit{Gist} features \cite{oliva2006building} to describe the visual content. 
The set of visual features are stored in the matrix $X$, and their corresponding textual descriptors are stored in the matrix $B$. 
Given a novel image feature (\textit{Gist}) $y$, and its noisy tag vector $h$, our goal is to obtain a refined estimate $\bar{h}$. 
We propose to exploit the correlations between the features and tags, using sparse coding, to perform tag refinement. 
Using the set of training examples, we construct the dictionary $\mathbf{D} = [X^T \text{ } \gamma B^T]^T$, 
where $\gamma$ is the scaling factor used to balance the total energy of features and tags. 
Similarly, the test sample is described as $z = [y^T \text{ } \gamma h^T]^T$. 
By assuming that the features and tags are clustered along subspaces, this structure can be discovered using sparse coding on examples:
\begin{eqnarray}
\min_{a} \|z - Da\|_2^2 + \lambda \ \|a\|_1 .
\label{eqn:joint_sc}
\end{eqnarray}The refined tag vector can then be estimated as $\bar{h} = B a$. The formulation in (\ref{eqn:joint_sc}) assumes that both features and semantic descriptors can be recovered using the same set of sparse coefficients. However, the $\ell_2$ penalty is not robust, and thus unsuitable for measuring the misfit in the reconstruction of tag vectors. 
To improve the recovery, we use different penalties for modeling visual features and tag vectors:
\begin{eqnarray}
\min_{a} \rho_1^{(1)}(y - X a) + \rho_1^{(2)}(h - B a) + \lambda \ \|a\|_1,
\label{eqn:joint_sc_mod}
\end{eqnarray}where $\rho_1^{(1)}$ is the $\ell_2$ penalty, and $\rho_1^{(2)}$ is the Huber penalty. 
As discussed in Section \ref{sec:alg_formulation}, the proposed framework can use the resultant mixed penalty to obtain sparse codes.

For our experiment, we used the Corel-5K data set \cite{duygulu2002object}, which is a very commonly used comparative data set for image annotation. 
There are $5,000$ images in total, and each image is annotated with $1$ to $5$ keywords. 
We used $4,500$ images as training data, and evaluated the performance using the rest. The total number of keywords in the vocabulary is $260$. 
We varied the level of noise in the test tags, by randomly flipping $\{1\%,3\%, 5\%, 10\%, 15\%, 20\%, 25\%\}$ of the entries in each binary tag vector. 
We estimated the refined tags using the both schemes described earlier, and computed the average noise ($\%$) in the refined tag vectors. 
Figure \ref{fig:tags}(a) plots the performance obtained using the $\ell_2$ penalty for the entire residual, and the mixed $\ell_2-$Huber penalty. 
As can be seen, the robust variant using the mixed penalty provides improved recovery at all noise levels. 
Furthermore, we corrupted the tag vectors of the training data also with different levels of noise and studied the performance deterioration (Figures \ref{fig:tags}(b) and (c)). 
We found that using mixed penalties provided superior performance in all cases.

\begin{figure*}[t]
\centering
\begin{subfigure}[b]{0.32\textwidth}    
 	\includegraphics[width=5.5cm]{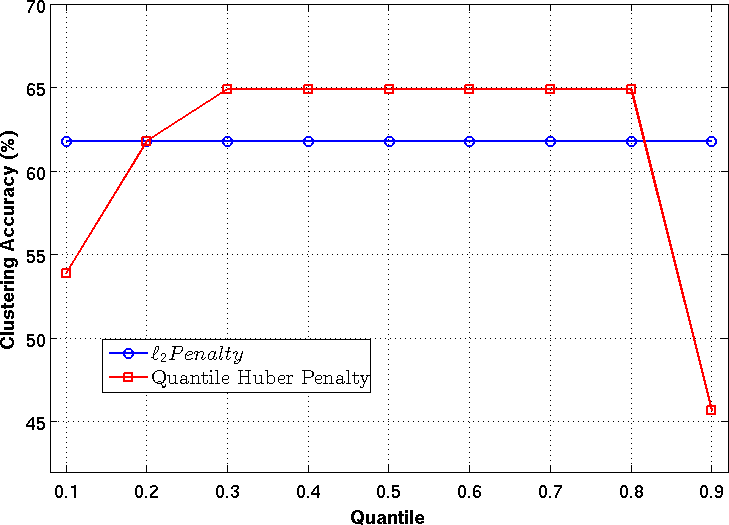}
 	\caption{Ecoli}           
\end{subfigure}%
\quad
\begin{subfigure}[b]{0.32\textwidth}
 	\includegraphics[width=5.5cm]{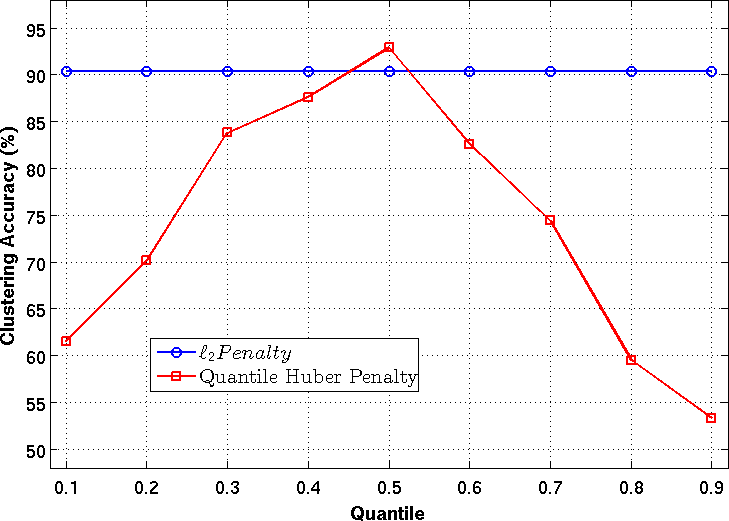}     
 	\caption{Wine}         
\end{subfigure}%
\quad
\begin{subfigure}[b]{0.32\textwidth}
 	\includegraphics[width=5.5cm]{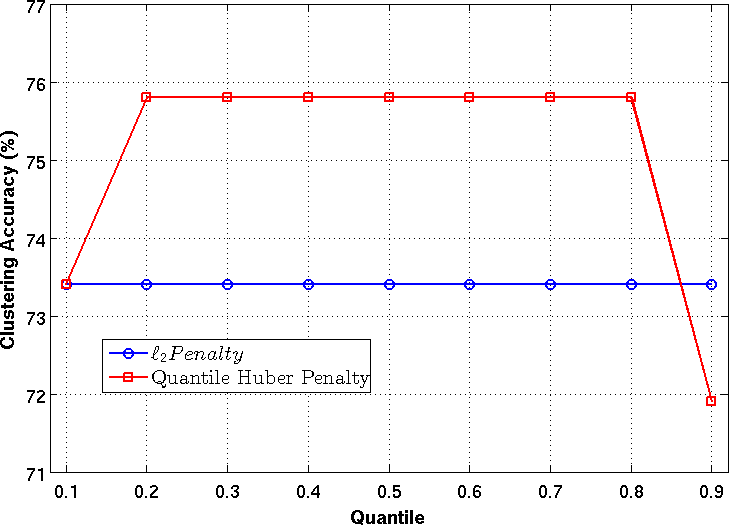}      
 	\caption{Breast Cancer}       
\end{subfigure}%
\caption{Evaluating the performance of subspace clustering using $\ell_1$-graphs at various quantiles for different datasets. The quantile-specific performances can be used to obtain empirical confidence limits on the median performance. The clustering performance with $\ell_2$ loss (blue) is also provided for comparison with quantile huber results (red).}
\label{fig:clus}
\end{figure*}

\subsection{Scenario 3: Computing Empirical Confidences for Subspace Clustering}
Assuming that the data samples lie in a union of subspaces allows us to perform unsupervised clustering using sparse coefficients \cite{Ramirez,l1graph}. 
By constructing a suitable graph to describe the relationship between data samples, we can analyze the eigen spectrum of the graph Laplacian to determine the underlying clusters. 
In particular, we can build an $\ell_1$ graph \cite{l1graph} for unlabeled data by solving for sparse codes using the data samples as the dictionary, 
with the constraint that a sample cannot contribute to its own representation. 
This procedure provides a non-local graph, as opposed to other locality-based graph construction strategies such as k-nearest neighbors. 
The coefficient matrix $A \in \mathbb{R}^{T \times T}$ from sparse coding is used to construct the graph Laplacian as $(\mathbb{I} - A)^T (\mathbb{I} - A)$, where $\mathbb{I}$ is the identity matrix. 
This model is very effective in several scenarios, but using the $\ell_2$ loss function to obtain the sparse code matrix $A$ makes it highly non-resilient to outliers. Furthermore, it is not possible to estimate confidence measures on the clustering performance using the $\ell_2$ loss function. We propose to employ the quantile Huber penalty to overcome these shortcomings, and generate reliable clusterings. 
Using the $\tau^{\text{th}}$ quantile to measure the misfit is equivalent to allowing a fraction $\tau$ of the entries in the residuals to be positive. 
Unless there is a complete model misfit, the penalty will deteriorate gradually as we consider quantiles away from the median. 
It was observed in the previous case studies that using the Huber ($0.5$ quantile) penalty makes sparse models robust to the outliers. 
In order to derive empirical confidence intervals and understand the reliability of clustering, we generate multiple $\ell_1$ graph based clusterings at different quantiles, 
and study their corresponding clustering performances.

In this experiment, we consider three datasets from the UCI repository: (i) Ecoli, (ii) Wine, and (iii) Breast cancer datasets. 
In each case, we build $\ell_1$ graphs at multiple quantiles between $0.1$ and $0.9$ and evaluate the clustering performance as the $\%$ Accuracy. 
Figure \ref{fig:clus} illustrates the clustering performances for the three datasets. 
In each case, we report the performance obtained by using the $\ell_2$ penalty for comparison. 
In addition to improving the clustering accuracy, using a more flexible loss function enables us to understand the reliability of the clustering results. 
For example, in case of the \textit{Wine} dataset (Figure \ref{fig:clus}(b)), though the median performance is high, the performance drops significantly as we move away from the median quantile. 
This shows that the clustering is very sensitive to the outliers, and small perturbations to the dataset might result in a sub-optimal performance. 
This behavior can be attributed to the limited availability of samples, or non-suitability of the chosen generative model. 
Though the union of subspace assumption seems valid for this data, lack of high confidence suggests that we choose different model assumptions for clustering.

\section{Conclusions}
\label{sec:concl}
Analyzing complex, high-dimensional data requires the design of interpretable, robust and scalable models. The proposed general framework has the inherent advantage of allowing one to compute sparse codes and optimize dictionaries using a broad class of misfit losses. However, the challenges in scaling this framework to large scale settings, and the inability to perform warm-starts with interior point methods make the design of online learning algorithms difficult. Using proximal methods, in lieu of interior point solvers, is a possible approach to overcome these shortcomings. In addition to enabling fast sparse code computation using warm-starts and online dictionary inference, this will allow us to go beyond sparsity regularization, and incorporate other penalties such as the nuclear norm (low-rank) regularization. Another important extension is to evaluate the proposed framework using graph penalties on the sparse codes for batch learning. It will be useful to incorporate similar penalties in online learning as well. We believe that such a general and scalable framework can expand the applicability of sparse models in data analytics. Some important application areas that can benefit by this development include matrix completion for recommender systems, topic modeling in text analytics, analysis of interactions in large networks, semantic content analysis in images/videos and data visualization.

\bibliographystyle{abbrv}
\bibliography{references_nrk,references_sasha}  


\end{document}